\pdfoutput=1

\documentclass[11pt]{article}

\usepackage[]{ACL2023}
\usepackage{multirow}

\usepackage{times}
\usepackage{latexsym}

\usepackage[T1]{fontenc}

\usepackage[utf8]{inputenc}
\usepackage{microtype}

\usepackage{inconsolata}

\usepackage{graphicx}
\usepackage{enumitem}
\usepackage{booktabs}
\usepackage{comment}
\usepackage{float}
\usepackage{pifont}
\newcommand{\cmark}{\ding{51}}%
\newcommand{\xmark}{\ding{55}}%

\newcommand{\datasetname}{PLAtE}
\newcommand{\fscore}{F1-score}
\newcommand{\mturk}{Mechanical Turk}

\title{\datasetname: A Large-scale Dataset for List Page Web Extraction}

\NewDocumentCommand{\heba}
{ mO{} }{\textcolor{purple}{\textsuperscript{\textit{Heba}}\textsf{\textbf{\small[#1]}}}}
\NewDocumentCommand{\kevin}
{ mO{} }{\textcolor{magenta}{\textsuperscript{\textit{Kevin}}\textsf{\textbf{\small[#1]}}}}
\NewDocumentCommand{\aidan}
{ mO{} }{\textcolor{cyan}{\textsuperscript{\textit{Aidan}}\textsf{\textbf{\small[#1]}}}}
\newcommand*\samethanks[1][\value{footnote}]{\footnotemark[#1]}

\author {
    Aidan San\thanks{\;\;The work was completed while Aidan and Yuan were interning at Amazon.}  \\ University of Virginia \\ \texttt{aws9xm@virginia.edu} \And 
    Yuan Zhuang\samethanks \\ University of Utah \\ \texttt{yuan.zhuang@utah.edu} \And 
    Jan Bakus\thanks{\;\;Jan left Amazon, but the work was completed while he was working at Amazon.} \\ Amazon  
    \AND
    Colin Lockard \\ Amazon \\ \texttt{clockard@amazon.com} \And
    David Ciemiewicz \\ Amazon \\ \texttt{ciemo@amazon.com} \And
    Sandeep Atluri \\ Amazon \\ 
 \texttt{satluri@amazon.com} \AND
    Yangfeng Ji \\ University of Virginia \\ \texttt{yangfeng@virginia.edu}\And
    Kevin Small \\ Amazon \\ \texttt{smakevin@amazon.com} \And
    Heba Elfardy \\ Amazon \\\texttt{helfardy@amazon.com}
}

\begin{document} 
\maketitle
\begin{abstract}
Recently, neural models have been leveraged to significantly improve the performance of information extraction from semi-structured websites. However, a barrier for continued progress is the small number of datasets large enough to train these models. In this work, we introduce the \datasetname{} (\textbf{P}ages of \textbf{L}ists \textbf{At}tribute \textbf{E}xtraction) benchmark dataset as a challenging new web extraction task. \datasetname{} focuses on shopping data, specifically extractions from product review pages with multiple items encompassing the tasks of: (1) finding product-list segmentation boundaries and (2) extracting attributes for each product. \datasetname{} is composed of $52,898$ items collected from $6,694$ pages and $156,014$ attributes, making it the first large-scale list page web extraction dataset. We use a multi-stage approach to collect and annotate the dataset and adapt three state-of-the-art web extraction models to the two tasks comparing their strengths and weaknesses both quantitatively and qualitatively. 
\end{abstract}
\section{Introduction}
\label{sec:introduction}
Semi-structured data extraction, i.e., web extraction, is the task of extracting data found in templated text fields from HTML pages. Once extracted, the structured data can be utilized in various downstream tasks such as information retrieval, recommendation, and question answering.

While recent work has shown the potential of neural approaches for web extraction \cite{lin2020freedom,zhou2021simplified,li2021markuplm}, 
there are very few publicly available large-scale datasets suitable for training and evaluation of these approaches, limiting progress in this area. 
Additionally, most existing datasets \cite{hao2011swde,hotti2021klarna} focus on one subset of the problem; namely \textit{detail page extraction}. In this paper, we introduce the \datasetname{} (\textbf{P}ages of \textbf{L}ists \textbf{At}tribute \textbf{E}xtraction) dataset\footnote{\datasetname{} will be publicly available at \texttt{\url{https://github.com/amazon-science/plate}}}, that specifically targets the task of \emph{list page} extraction  and focuses on product review pages in the shopping vertical, i.e., multi-product review pages. 

To elaborate, item pages can be broadly categorized into two classes: \textit{detail pages} and \textit{list pages}. Detail pages provide detailed information about a single item. List pages comprise a list of items with abridged detail, organized under a single theme, e.g. ``best board games''. This organization facilitates direct comparison of each item and allows for the extracted data to be easily integrated into recommender, question answering, or dialogue systems powering digital assistants \cite{linden2003recommendations, gupta2019amazonqa, zhang2018saur}. Extracted product data can be utilized by both content creators (publishers) as well as customers looking to make purchase decisions. \footnote{We provide an example of a list page and a detail page in the appendix.} 


\begin{figure}
\centering
\includegraphics[width=.7\columnwidth]{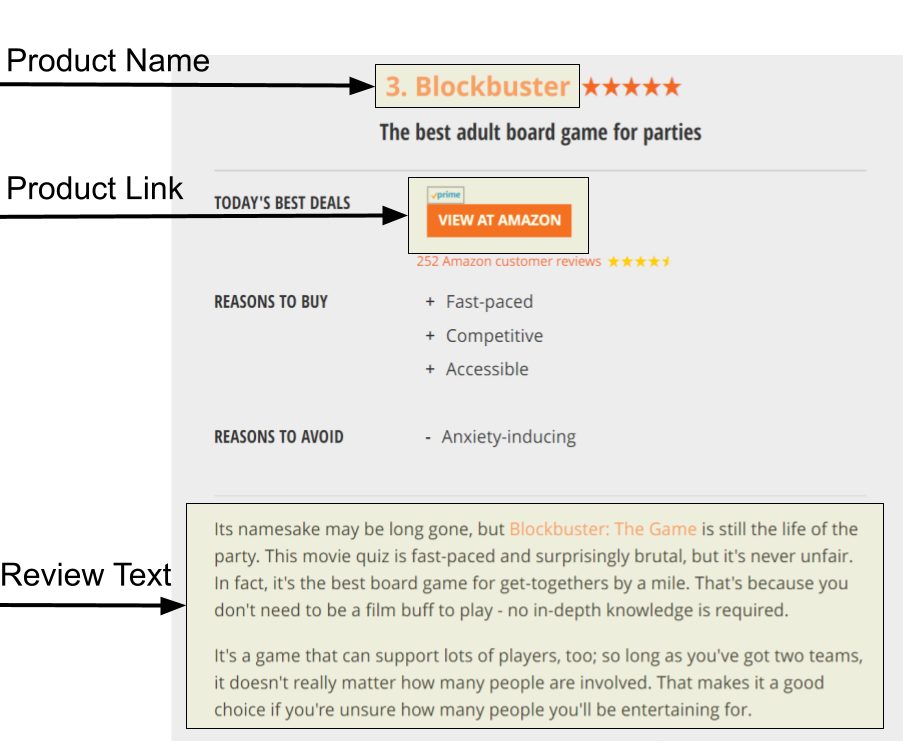}
\caption{A single item (i.e. product) from a list page for a board game (from {\tt gamesradar.com}).}
\label{fig:schema}
\vspace{-5mm}
\end{figure}

Because \datasetname{} is built from list (multi-item) pages, we can evaluate two tasks: segmentation and attribute extraction. In the segmentation task, the model should determine the boundaries of each product, i.e. where the information for each product begins and ends. In the attribute extraction task, each node in the HTML DOM tree can be assigned any number of three labels: product name, product review text, and product link. 
The dataset is comprised of $52,898$ products from $6,694$ pages split into: training, development, and held-out test sets.
To evaluate the dataset, we adapt two state-of-the-art neural web extraction models;  MarkupLM \cite{li2021markuplm} and DOM-LM \cite{deng2022dom}, as well as explore a text-based Transformer model, RoBERTa \cite{roberta}.
We achieve an \fscore{} of $0.787$ and $0.735$ for segmentation and attribute extraction tasks respectively. We evaluate the potential of multi-task learning to improve performance and find that multi-task learning improves recall but slightly decreases precision, for both tasks. 
To summarize, our contributions are: (1) creating the first large-scale \textit{list page} web extraction dataset, (2) adapting state-of-the-art  neural web extraction approaches to the segmentation and attribute extraction tasks, and (3) qualitatively and quantitatively comparing the performance of different models on the two presented tasks; segmentation and attribute extraction.


\section{Related Work}
\label{sec:relatedwork}
\begin{table*}
\centering
\small
\begin{tabular}{lccccccc}
\toprule
\textbf{Name} & \textbf{Pages/Vert.} & \textbf{Sites/Vert.} &  \textbf{Records/Vert.} & \textbf{Tot. Pages} & \textbf{Year} & \textbf{Mult. Item?} \\
\midrule
SWDE \cite{hao2011swde} & 4,405-20,000 & 10 & 4,405-20,000 & \underline{124,291} & 2011 & \xmark \\
WDC \cite{conf:ecweb:PetrovskiPMB16} & 576 & ? & 576 & 576 & 2016 & \xmark \\
Klarna \cite{hotti2021klarna} & \underline{51,701} & \underline{8,175} & 51,701 &  51,701 & 2019 & \xmark \\
LDST \cite{zhu2006simul} & 771 & ? & $\sim8600$ & 771  & 2006 & \cmark \\ 
\cite{dhillon2011mtlearning} & 5-15 & 5-8 & $\sim400$ & $\sim30$ & 2011 & \cmark \\
AMBER \cite{furche2012amber} & 150-281 & \textbf{100-150} & 1608-2785 & 431 & 2012 & \cmark \\
\datasetname{} & \textbf{6,694} & 43 & \underline{\textbf{52,898}} &  \textbf{6,694}
 & \underline{\textbf{2020}} & \cmark \\
\bottomrule
\end{tabular}
\caption{Comparison of existing web extraction datasets against \datasetname{}. \datasetname{} has the greatest number of records per vertical, and the freshest HTML content. Additionally, it has an order of magnitude larger number of records than any of the other multiple item datasets. ``?'' Means the information is not available in the paper. The multi-item dataset with the highest statistic is bolded and the overall dataset with the highest statistic is underlined. 
}
\label{tab:dscomparison}
\end{table*}

The vast majority of previous web extraction datasets (e.g., SWDE ~\cite{hao2011swde}, the Klarna Product Page Dataset \cite{hotti2021klarna}, and WDC \cite{conf:ecweb:PetrovskiPMB16}) are composed of single item pages. 
Multiple item or list page datasets are much less common. \citet{zhu2006simul} created a dataset of 771 list pages,  while \citet{dhillon2011mtlearning} created a small dataset (BSCM) of about 30 pages from 4 verticals: Books, Seminars, CS Faculty and MLConfs. \citet{furche2012amber} built a dataset of 431 pages from two verticals: UK used car dealer websites collected from a used car aggregator website and UK real-estate websites collected from the yellow pages.
To the best of our knowledge, \datasetname{} is the largest multi-item web extraction dataset.  Additionally, most previous web extraction datasets assume a single DOM node has at most a single label. However, this assumption does not hold true in many product pages. For example, for many products, the product name's text-span is also a link to the product. Our dataset does not have this assumption; we allow a node to have multiple labels. Finally, most existing list page datasets are not publicly available. Table \ref{tab:dscomparison} compares the different web extraction datasets. 

From the methods side, web extraction has first been tackled using wrapper induction methods that create a set of rules (wrappers) to transform unstructured input into structured output \cite{kushmerick1997wrapper,furche2014diadem, zheng2007jointwrapper,Azir2017WrapperAF,gulhane2011vertex,carlson2008bootstrapping,furche2012amber}.
Recently, a number of advances have been made by utilizing neural-based approaches to construct a representation for each HTML node for the extraction task \cite{lockard2020zeroshotceres, lin2020freedom, zhou2021simplified,li2021markuplm,deng2022dom,xie2021webke}.

Other tasks related to semi-structured information extraction include boilerplate removal \cite{leonhardt2020boilerplate},  
extraction from HTML tables \cite{Cafarella2018TenYO,deng2020turl,wang2021tcn,herzig-etal-2020-tapas}, 
and segmenting pages, e.g., using clustering followed by string
alignment \cite{alvarez2008}, optimization based on divide and
conquer \cite{bing2013robust}, and Hierarchical Conditional Random Fields (CRFs) \cite{zhu2006simul}.



\section{\datasetname{} Benchmark}
\label{sec:benchmark}
In this work, we tackle two tasks: {\it (1) segmentation}, i.e., identifying the boundaries of the individual products in a given page and {\it  (2) attribute extraction}, i.e., identifying the individual attributes for each identified product. For each given product, we extract the following three attributes:
%
(1) \textit{product name}: This refers to the name of the product, e.g. ``iPhone 11'', (2)
\textit{review text}: This is generally a high-level review or general description of the product, and
(3) \textit{purchase link}: a link (or button) to a merchant's website (e.g., Amazon, Ebay). We commonly see generic text such as ``Buy Here'', the name of the merchant such as ``Amazon'', or the name of the product. 
%
%
Similar to prior work, we perform classification on the leaf nodes; i.e., only nodes that contain text are passed to the classification models. 
\subsection{\datasetname{} Construction Process}

To construct \datasetname{}, we started with $270$M documents from the large, publicly available web crawl Common Crawl.\footnote{\tt https://commoncrawl.org/} We then filtered down to $6,694$ candidate pages from $43$ internet websites by (1) removing duplicate URLs and non-English pages, (2) filtering out non-multi-product pages using a linear classifier with word embedding and TF-IDF features as well as keywords-based heuristics (e.g., ``best'', ``compare'', etc.), (3) selecting top/popular websites using the Tranco List \cite{Pochat2019TrancoAR},  (4) selecting sites with the highest number of list pages, and (5) filtering out pages with inappropriate (i.e. sensitive or offensive) content. 

After selecting the candidate pages, we performed the initial segmentation and attribute extraction by using CSS selector rules.\footnote{CSS selector rules are patterns composed of HTML tag names and HTML class names used to select one or more  HTML elements (e.g., 
{\tt [.product-slide p]})} 
Two expert annotators used a custom annotation tool to annotate a representative page from each site by selecting a CSS selector for each attribute. 
The annotated CSS selectors were then used to extract the attributes from the rest of the pages from the same site.
Multiple rounds of quality checks were performed in order to ensure the quality of the final selector rules.

\begin{figure*}
\centering
\includegraphics[width=.80\linewidth]{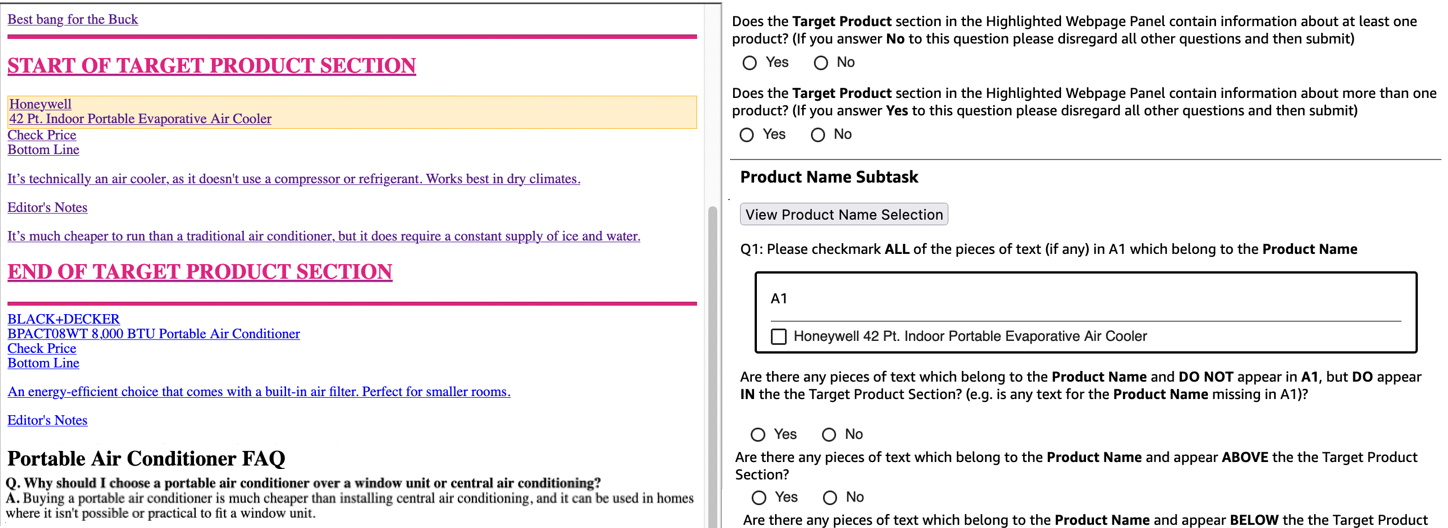}
\caption{Sample \mturk{} task. In the left panel, the text-span ``Honeywell42.  Pt Indoor Portable Evaporative Air Cooler'' is highlighted as the extraction of interest. In the right panel, a checkbox denotes whether the highlighted text-span falls under the named attribute, i.e., ``Product Name''. The annotator is also asked to determine whether the left panel is a valid product and if any named attributes are missing.}
\label{fig:mturk}
\end{figure*}

The final step in creating \datasetname{} used Amazon \mturk\ annotations in order to remove any errors introduced by the rule-based extraction step. For the \mturk{} annotation task, we presented a web page to the annotators. We first asked the annotators a set of page-level questions to ensure that the web page is a valid multi-product review page. We then asked the annotators to verify that a piece of highlighted text within the web page should be extracted as an attribute and asked them to indicate if any text of the attribute of interest was not highlighted, i.e., was not captured by the rules.\footnote{We only qualified annotators from English speaking countries that completed at least $100$ prior annotation tasks with an acceptance rate of $95\%$ or more, and who passed a custom qualification task.} Overall, $20\%$ of workers that attempted the qualification task were qualified, resulting in $77$ annotators. 
To identify spammers, we blocked any annotator that spent less than $20$ seconds on average per annotation task. Finally, to minimize annotation costs while ensuring high-quality evaluation and development data, we used one annotation per task for the training set and three annotations per task for the development and test sets. Majority vote was used to aggregate the annotations from the development and held-out test sets.

To build the final dataset, we split the data such that the training, development, and held-out test sets have approximately the same distribution in terms of number of products and pages. Moreover, we ensured that sites from the same website family, e.g., \emph{thespruce.com} and \emph{thespuceeats.com}, appear in the same split. 
Table \ref{tab:dsexample} shows a sample \datasetname{} annotation, while Table \ref{tab:dssize} shows statistics of the dataset.
\begin{table}
\small
\centering
\begin{tabular}{ll}
\toprule
Site & theinventory.com \\
URL & theinventory.com/best-iphone ... \\
Product Index & 4 \\
Attr. Name & Product Name \\ 
Num. Extracted & 1 \\
XPath & /html/body/div[3]/ ... /span/a/strong  \\ 
Text & ['AirPods Pro'] \\
\bottomrule
\end{tabular}
\caption{An example \datasetname{} annotation.} 
\label{tab:dsexample}
\end{table}

\begin{table}
\small
\centering
\begin{tabular}{lrrrr}
\toprule
\textbf{Split} & \textbf{\# Sites} & \textbf{\# Pages} & \textbf{\# Products} & \textbf{\# Attrs} \\
\midrule
Train & $28$ & $4,202$ & $35,383$ & $103,731$\\
Dev & $5$ & $655$ & $6,038$& $18,019$ \\
Test & $10$ & $1,837$ & $11,477$ & $34,264$\\
\midrule
All & $43$ & $6,694$ & $52,898$ & $156,014$\\
\bottomrule
\end{tabular}
\caption{Statistics of the train, development, and test sets.}
\label{tab:dssize}
\end{table}

\begin{table*}
\centering
\small
\begin{tabular}{llp{1\columnwidth}cl}
\toprule
\textbf{Attribute} & \textbf{Tag} &\textbf{Text} & \textbf{Crowd Annotations} & \textbf{Gold}\\
\midrule
Review Text & \texttt{<p>} & And those are our recommendations for the best mattresses!...
& [True, False, False]  & False \\
Product Link & \texttt{<a>} & Tempur-Pedic & [True, True, False] & True\\
Review Text & \texttt{<p>} & And those are our recommendations for best outdoor grills...  
& [True, True, False]  & False \\
Product Link & \texttt{<a>} & Original Sprout's miracle detangler & [True, False, False] & True\\
\bottomrule
\end{tabular}
\caption{Annotator disagreements on the page {\tt https://www.wisebread.com/the-5-best-mattresses}.}
\label{tab:turkdisagree}
\end{table*}

%
\label{sec:dataanalysis}
\subsection{Dataset Analysis}
\begin{table}
\setlength{\tabcolsep}{3.5pt}
\centering
\small
\begin{tabular}{lcccc}
\toprule
\textbf{Attribute} & \textbf{Missing(\%)} & \textbf{Valid (\%)}& \textbf{Fleiss-Kappa} \\
\midrule
Product Name   & $2.7$ & $97.0$ & $0.596$\\
Review Text    & $2.3$  & $85.1$ & $0.728$\\
Product Link   & $6.0$ & $96.7$ & $0.560$\\
\bottomrule
\end{tabular}
\caption{Annotation statistics.} 
\label{tab:mturk}
\end{table}
We manually analyzed 
\datasetname{} annotations to assess their quality.
In the annotation task, as seen in Figure \ref{fig:mturk}, the first two annotation questions were designed to filter out cases of mislabeled or malformed snippets.
Specifically, the first question asked whether the annotation snippet contained at least one product while the second question asked if the snippet contained more than one product. Annotators indicated that $99.6\%$ of snippets contained at least one product and that only $0.8\%$ contained more than one product, indicating that the vast majority of tasks sent to annotators were valid product snippets.
The next set of annotation questions are meant to identify if any annotations were missing from the presented snippet. Between $3\%$ and $6$\% of the extracted attributes were missing some text spans. Finally, the last set of questions asked the annotators to select the text-spans that matched an attribute from a set of check-boxes (all text-spans matching the CSS rule). Overall, more than $85\%$ of the text-spans were correctly matched to the corresponding attribute. 
We also calculated the inter-annotator agreement using Fleiss-Kappa \cite{Fleiss1973TheEO} to measure the quality of the annotation guidelines and found that the inter-annotator agreement is moderate for ``\emph{Product Name}'' and ``\emph{Product Link}'' and substantial for ``\emph{Review Text}''.
For ``\emph{Product Link}'', some product name spans were also product link spans, causing confusion among the raters. The annotation statistics are shown in Table \ref{tab:mturk}. Percentage of missing attributes is relatively low indicating that our CSS selector rules have high recall. Additionally, percentage of valid extractions is quite high especially for \textit{Product Name} and \textit{Product Link} indicating that our CSS selector rules have high precision. 
Table \ref{tab:turkdisagree} shows examples of annotator disagreement. The third example is challenging because the text is about outdoor grills in general, and not a particular outdoor grill product, so it should be annotated as False. In the fourth example, the annotators likely missed the correct label (\emph{True}), because the link does not follow the standard format of ``\emph{Buy Now}'' or a retailer's name such as ``\emph{Amazon}''.

\section{Models}

\label{sec:results}
\begin{table*}
\centering
\small
\begin{tabular}{l|ccc|ccccc}
\toprule
&\multicolumn{3}{c|}{\textbf{Attribute Extraction}} & \multicolumn{5}{c}{\textbf{Segmentation}}\\
\textbf{Model} & \textbf{Test P} & \textbf{Test R} & \textbf{Test F1} & \textbf{Test P} & \textbf{Test R} & \textbf{Test F1} & \textbf{Test ARI} & \textbf{Test NMI} \\
\midrule
RoBERTa &\underline{$0.843$} & $0.652$ & \underline{$0.735$} & $0.692$ & $0.665$ & $0.678$ & $0.693$ & $0.744$\\
DOMLM & $0.815$ & \underline{$0.655$} & $0.726$ & $0.718$ & $0.728$ & $0.722$ & $0.716$ & $0.764$ \\
MarkupLM & $0.839$ & $0.620$ & $0.711$ & \underline{$0.769$} & \underline{$0.805$} & \underline{$0.787$} & \underline{$0.771$} & \underline{$0.870$}\\
\bottomrule
\end{tabular}
\caption{Performance of the different models on the segmentation and attribute extraction tasks. For segmentation, MarkupLM has the best performance while for attribute extraction, DOM-LM outperforms RoBERTa on Recall, but RoBERTa overall has the highest F1. }
\label{tab:attrextr}
\end{table*}

\label{sec:model}
We evaluate the performance of three recent neural models: RoBERTa \cite{roberta}, DOM-LM \cite{deng2022dom} and MarkupLM \cite{li2021markuplm} on \datasetname{}. 
\paragraph{RoBERTa}
is a Transformer-based~\cite{vaswani2017attention} model pre-trained with natural language texts sourced from the BookCorpus \cite{zhu2015aligning} and Wikipedia. The pre-training task is masked language modeling. In our experiments, the input to RoBERTa is a sequence of text tokens from the DOM tree. Consequently, it does not utilize other types of information from the DOM tree such as XPaths or HTML tags.      
\paragraph{DOM-LM}
is designed to generate contextualized representations for HTML documents. It uses RoBERTa as the base encoder and is pre-trained over the SWDE dataset~\cite{hao2011swde} with masked language modeling over the text tokens as well as the DOM tree nodes. To encode a DOM tree, DOM-LM first slices the tree into a set of subtrees such that important tree-level context is kept in each subtree. Then each  node is represented by its tag, text, class/id attributes, as well as a positional matrix based on its position in the DOM tree. 

\paragraph{MarkupLM}
is another RoBERTa-based model. The input to MarkupLM is a node represented by both an XPath embedding and its text. The XPath embedding is created by embedding each tag and subscript in the XPath separately and then concatenating and passing them through a FFN layer. The model was pre-trained on three tasks: (1) masked markup language modeling, (2) node relation prediction, and (3) title page matching using 24 million pages from Common Crawl.
%

For the segmentation task, we label each of the nodes as \textit{begin} (B) to denote the first text-span of a product, or \textit{other} (O) for the rest of the nodes.
We then apply a softmax layer to the logits and train with cross-entropy loss.
For the attribute extraction task, the model predicts the attribute labels from the logits using a multi-label sigmoid layer that was trained with binary cross-entropy loss. As a multi-label classification task, the model can assign each node with any subset of the labels or no label.
In both tasks, we begin with one of the three pre-trained models and then fine-tune on the training set of \datasetname{}. 

\section{Experiments}
For both segmentation and attribute extraction tasks, we report the precision, recall, and \fscore{}. Results are macro-averaged over the different classes. In addition, for segmentation, we report clustering metrics following \cite{bing2014wpsgraph}, where the attributes in the same segment are considered to be in the same cluster. In our dataset, when looking at two adjacent products in a page, there is a single node which we label as a ``segmentation boundary''. In reality, there can be multiple nodes which appear between two adjacent products and split the products apart. If multiple valid segmentation boundaries for a product are possible, \fscore{} will penalize the model for picking any segmentation boundary which is not labelled as such. On the other hand, the clustering metrics provide a relaxation which checks that product attributes are assigned to the correct product. For clustering metrics, we report adjusted rand index (ARI) \cite{Hubert1985ARI} and normalized mutual information (NMI) \cite{Strehl2002NMI} where a higher number indicates better performance. \footnote{ARI ranges from $-0.5$ to $1$ and NMI ranges from $0$ to $1$.}\textsuperscript{,}\footnote{We average all results for each of the two tasks over three runs from different random seeds.} 

Overall, we observe that MarkupLM performs well on the segmentation task yielding scores of $0.787$, and $0.771$ and $0.870$ 
for F1, ARI and NMI respectively while DOM-LM performs worse on this task with scores of $0.722$ and $0.716$ and $0.764$ for F1, ARI and NMI respectively. This can likely be attributed to the fact that MarkupLM was pre-trained on the task of relation prediction between nodes. Identifying if two nodes have a parent-child, sibling, or ancestor-descendant relation could help the model distinguish nodes within the same product from nodes of different products, and consequently identify the product boundaries better.



\begin{table}
\centering
\small
\begin{tabular}{llccc}
\toprule
\textbf{Model} & \textbf{Attribute} & \textbf{P} & \textbf{R} & \textbf{F1}\\
\midrule
 & Product Name & $0.858$ & $0.645$ & $0.737$ \\
RoBERTa & Review Text & \underline{$0.922$} & \underline{$0.721$} & \underline{$0.808$} \\
 & Product Link & $0.750$ & $0.590$ & $0.661$ \\
\midrule
 & Product Name & $0.843$ & \underline{$0.672$} & \underline{$0.747$} \\
DOM-LM & Review Text & $0.863$ & $0.695$ & $0.769$ \\
& Product Link & $0.741$ & \underline{$0.599$} & $0.662$ \\
\midrule
 & Product Name & \underline{$0.885$} & $0.621$ & $0.730$ \\
MarkupLM & Review Text & $0.822$ & $0.671$ & $0.737$ \\
 & Product Link & \underline{$0.808$} & $0.569$ & \underline{$0.667$} \\
\bottomrule
\end{tabular}
\caption{Precision, recall, and \fscore{} for 
all models on the attribute extraction task by attribute type.} 
\label{tab:attrextrbylabel}
\end{table}
The attribute extraction task shows a different trend from segmentation where RoBERTa performs the best, outperforming MarkupLM and DOM-LM. We look into the reason behind this in Section \ref{sec:discussion}. Detailed results for both tasks are presented in Table \ref{tab:attrextr} while precision, recall, and \fscore{} for the attribute extraction task broken down by attribute are shown in Table \ref{tab:attrextrbylabel}. We find that product link extraction performs  worst while product review text performs best. 
For product link, this difficulty can be attributed to the diverse nature of product links which can take the form of a \emph{product name}, or the \emph{price of the product} e.g., a hyperlink with the text $\$10$ or a \emph{button} (e.g. with the text ``Buy Now''). For review text, the higher performance is not surprising given that it normally has a more consistent style than product links.

To better understand the different factors affecting the performance of the attribute extraction task, we (1) break down the scores by site in our test set to see whether some sites are more challenging than others, (2) analyze the relationship between the number of products in a page and attribute extraction performance (3) explore whether the index of the product (i.e., where it appears in the page) affects the performance, (4) study the effect of adding more pages (versus sites) to the training data, and (5) explore whether a multi-task model that jointly tackles both segmentation and attribute extraction tasks can yield better performance through utilizing the complementary signals between both tasks.  
%
\begin{figure}
    \centering
    \includegraphics[width=0.7\columnwidth]{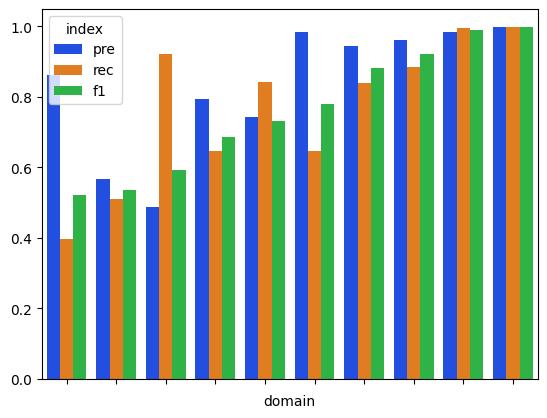}
    \caption{Precision, recall, and F1 of MarkupLM's attribute extraction scores grouped by site. We observe a high variance in scores, due to differing HTML structure between different sites. 
}
\label{fig:scorebysite}
\end{figure}

As can be seen in Figure \ref{fig:scorebysite}, extraction performance varies greatly by site. We suspect that this is due to the diverse page layouts and styles of different sites. When grouping pages by number of products, we find that for all three models, as the number of products increases, the \fscore{} improves. We believe this is due to the increased ease of the model to learn patterns within a page given more examples in a page. When grouping the scores by product index, we find that product index does not have a significant impact on the performance of attribute extraction; i.e. performance is relatively uniform across different locations in the page.\footnote{Figures showing the detailed results of these experiments are provided in the appendix.} 

%
To study the value of annotating more sites compared to annotating more pages, we sampled two subsets of pages from our original training set. We collected all pages from $14$ randomly sampled sites from the training set for a total of $1514$ pages, then sampled $1514$ random pages across all sites in the training set. The $14$ sites achieved an F1=$0.595$ compared to sampling from all sites achieving an F1=$0.681$. From this, we can conclude there is indeed value in annotating a greater sites instead of simply annotating more pages from the same site. We suspect the model is better able to generalize to the test set, when provided with a more diverse training set of different sites. Finally, we look into whether multi-task learning improves model performance on \datasetname{}. To this end, we train a MarkupLM model with a shared encoder, but distinct loss functions and output layers for the segmentation and attribute extraction tasks.\footnote{We use a weighted sum of the loss functions for both tasks and determine the weights empirically based on the performance on the development set.}
We find that multi-task learning improves the recall of both tasks -- as shown in Table \ref{tab:multitask} -- but slightly harms precision. \fscore{} performance increases by $0.023$ 
for the segmentation task but goes down for the attribute extraction task due to the decrease in precision.



\begin{table}
\centering
\small
\begin{tabular}{lcccc}
\toprule
\textbf{Task} & \textbf{Setting} & \textbf{P} & \textbf{R} & \textbf{F1} \\
\midrule
\multirow{2}{1.2cm}{Attribute Extraction} & No MTL & \underline{$0.839$} & $0.620$ & \underline{$0.711$} \\
&  MTL \scriptsize{(weight=0.5)} & $0.803$ & \underline{$0.628$} & $0.703$\\
\midrule
\multirow{2}{1.34cm}{Segmentation} & No MTL & \underline{$0.769$} & $0.805$ & $0.787$ \\
 & MTL \scriptsize{(weight=0.25)} & $0.768$ & \underline{$0.859$} & \underline{$0.810$}\\
\bottomrule
\end{tabular}
\caption{Precision, recall, and F1 for Multitask Learning (MTL) for the MarkupLM model.}
\label{tab:multitask}
\end{table}



\section{Discussion}
\label{sec:discussion}
Contrary to our expectations, RoBERTa outperformed DOM-LM, a model specially designed to model webpages, on the attribute extraction task. As can be seen in Table \ref{tab:attrextrbylabel}, RoBERTa mainly outperforms DOM-LM in review text precision. We analyze 20 examples where DOM-LM makes a false positive review text error and RoBERTa does not. We find in 95\% of the examples, the misclassified node is outside of the main review content, and in 80\% of the examples the misclassified node is a  \texttt{<p>} tag. This indicates that DOM-LM's structural encoder is likely over-fitting to the HTML tag and disregarding the text content, hence is less able to generalize to unseen HTML structures.

We perform additional analysis of the outputs from the MarkupLM model, to identify areas for improvement for the attribute extraction task. Specifically, we sampled up to $5$ false positive (FP) and $5$ false negative error examples for each attribute from each site in the test set, and ended up with a total of $257$ examples.\footnote{Some sites had less than $5$ errors, in which case we examined all errors.}
For product review text, we find that many of the false negatives are very short textspans such as ``\emph{These}'', ``\emph{as well}'', and ``\emph{require}'' representing a single text leaf node within a larger paragraph. We suspect that the model has not been trained with enough examples of varied text length. In the future, we could also consider training the model to classify the parent nodes representing a whole paragraph instead of classifying at the leaf node level.
For false positives, we find $4$ examples of user-written comments (as opposed to reviewer-written text). While these do not represent the official review text, they are semantically similar to the official review text hence are easily confused by the model. One such example is a user's comment mentioning that ``\emph{the design and functionality of these cookers is top-notch}'' which has a similar style to text which could have been written by a reviewer. For product link, $17$ false positives are hyperlinks that link to a homepage rather than to a specific product which indicates that the model is over-fitting to the \texttt{<a>} tag. Training the model using contrastive learning with positive and negative product link \texttt{<a>} tag examples should help with this case. \\ 
%

Next we explore the effect of threshold-tuning on the performance of attribute extraction.  In the presence of an oracle that specifies whether or not a node should be tagged with an attribute, we can use the $argmax$ of the logits of the different attribute classes in order to guarantee that the node is tagged with one of the three attributes. Based on the very high agreement between the $argmax$ and the actual label for product name (89.4\%) and product review text (91.8\%), it is clear that some of the challenges in tagging nodes in PLAtE stem from needing to determine whether or not a node should be tagged with any attribute, on top of determining what the correct attribute tag is.


Finally, we analyze the errors in the segmentation task, collecting all pages with an ARI and NMI $<0.5$ within test set. We find that in 97\% of these pages the number of gold segmentation boundaries is higher than the number of predicted segmentation boundaries. This means that when ARI and NMI are very low, the model is failing to split the page into enough segments. To improve performance, we could consider utilizing a structured prediction to model segmentation interdependencies e.g. the model could explicitly model that it is unlikely that two segmentation boundaries appear directly next to one another. 

\section{Conclusion}
\label{sec:conclusion}
In this work, we introduce \datasetname{}, the first large-scale list page dataset for web extraction. We describe our methodology in creating the dataset, from pre-processing to crowd-sourced annotation collection. We evaluate the performance of three strong web extraction baselines and achieve an \fscore{} of $0.787$ and $0.735$ on the segmentation and attribute extraction tasks respectively. 
While we focus on shopping domain due to its importance to several downstream applications, in the future we intend to extend our work to other verticals to facilitate further research studying model generalization and domain adaptation. 

\section*{Acknowledgements}
We would like to thank the anonymous reviewers for their detailed and insightful comments and suggestions. 
We would also like to thank Wendy Wei, Markus Dreyer, Polly Allen, Yusuke Watanabe, Matthias Petri, Tian Tang, Phi Nguyen, Ritesh Sarkhel, Hengxin Fun, Mengwen Liu, and Duke Vijitbenjaronk for their suggestions and assistance with this work.

\section*{Limitations}
To ensure high quality extractions for \datasetname{}, we optimize our annotation process for precision. For example, for the Product Link attribute, we generally annotate only one product link per product. In an application scenario, the user would not need multiple links to a purchase page, but this could potentially harm the precision of the evaluated models. 
In addition, we assume that all attributes are text-based. This has the potential of missing additional product information which could be helpful to users, such as images of the product. In future work, we would like to extend \datasetname{} by incorporating other modalities.

\bibliography{anthology,custom}

\begin{thebibliography}{36}
\expandafter\ifx\csname natexlab\endcsname\relax\def\natexlab#1{#1}\fi

\bibitem[{Azir and Ahmad(2017)}]{Azir2017WrapperAF}
Mohd Azir and Kamsuriah Ahmad. 2017.
\newblock Wrapper approaches for web data extraction : A review.
\newblock \emph{2017 6th International Conference on Electrical Engineering and
  Informatics (ICEEI)}, pages 1--6.

\bibitem[{Bing et~al.(2014)Bing, Guo, Lam, Niu, and Wang}]{bing2014wpsgraph}
Lidong Bing, Rui Guo, Wai Lam, Zheng-Yu Niu, and Haifeng Wang. 2014.
\newblock \href {https://doi.org/10.1145/2600428.2609630} {Web page
  segmentation with structured prediction and its application in web page
  classification}.
\newblock In \emph{Proceedings of the 37th International ACM SIGIR Conference
  on Research \& Development in Information Retrieval}, SIGIR '14, page
  767–776, New York, NY, USA. Association for Computing Machinery.

\bibitem[{Bing et~al.(2013)Bing, Lam, and Wong}]{bing2013robust}
Lidong Bing, Wai Lam, and Tak-Lam Wong. 2013.
\newblock \href {https://doi.org/10.1145/2508434} {Robust detection of
  semi-structured web records using a dom structure-knowledge-driven model}.
\newblock \emph{ACM Trans. Web}, 7(4).

\bibitem[{Cafarella et~al.(2018)Cafarella, Halevy, Lee, Madhavan, Yu, Wang, and
  Wu}]{Cafarella2018TenYO}
Michael~J. Cafarella, Alon~Y. Halevy, Hongrae Lee, Jayant Madhavan, Cong Yu,
  Daisy~Zhe Wang, and Eugene Wu. 2018.
\newblock Ten years of webtables.
\newblock \emph{Proc. VLDB Endow.}, 11:2140--2149.

\bibitem[{Carlson and Schafer(2008)}]{carlson2008bootstrapping}
Andrew Carlson and Charles Schafer. 2008.
\newblock Bootstrapping information extraction from semi-structured web pages.
\newblock In \emph{Proceedings of the 2008th European Conference on Machine
  Learning and Knowledge Discovery in Databases - Volume Part I}, ECMLPKDD'08,
  page 195–210, Berlin, Heidelberg. Springer-Verlag.

\bibitem[{Deng et~al.(2022)Deng, Shiralkar, Lockard, Huang, and
  Sun}]{deng2022dom}
Xiang Deng, Prashant Shiralkar, Colin Lockard, Binxuan Huang, and Huan Sun.
  2022.
\newblock Dom-lm: Learning generalizable representations for html documents.
\newblock \emph{arXiv preprint arXiv:2201.10608}.

\bibitem[{Deng et~al.(2020)Deng, Sun, Lees, Wu, and Yu}]{deng2020turl}
Xiang Deng, Huan Sun, Alyssa Lees, You Wu, and Cong Yu. 2020.
\newblock \href {http://arxiv.org/abs/2006.14806} {Turl: Table understanding
  through representation learning}.

\bibitem[{Dhillon et~al.(2011)Dhillon, Sellamanickam, and
  Selvaraj}]{dhillon2011mtlearning}
Paramveer~S. Dhillon, Sundararajan Sellamanickam, and Sathiya~Keerthi Selvaraj.
  2011.
\newblock \href {https://doi.org/10.1145/2063576.2063713} {Semi-supervised
  multi-task learning of structured prediction models for web information
  extraction}.
\newblock In \emph{Proceedings of the 20th ACM International Conference on
  Information and Knowledge Management}, CIKM '11, page 957–966, New York,
  NY, USA. Association for Computing Machinery.

\bibitem[{Fleiss and Cohen(1973)}]{Fleiss1973TheEO}
Joseph~L. Fleiss and Jacob Cohen. 1973.
\newblock The equivalence of weighted kappa and the intraclass correlation
  coefficient as measures of reliability.
\newblock \emph{Educational and Psychological Measurement}, 33:613 -- 619.

\bibitem[{Furche et~al.(2014)Furche, Gottlob, Grasso, Guo, Orsi, Schallhart,
  and Wang}]{furche2014diadem}
Tim Furche, Georg Gottlob, Giovanni Grasso, Xiaonan Guo, Giorgio Orsi,
  Christian Schallhart, and Cheng Wang. 2014.
\newblock \href {https://doi.org/10.14778/2733085.2733091} {Diadem: Thousands
  of websites to a single database}.
\newblock \emph{Proc. VLDB Endow.}, 7(14):1845–1856.

\bibitem[{Furche et~al.(2012)Furche, Gottlob, Grasso, Orsi, Schallhart, and
  Wang}]{furche2012amber}
Tim Furche, Georg Gottlob, Giovanni Grasso, Giorgio Orsi, Christian Schallhart,
  and Cheng Wang. 2012.
\newblock \href {https://doi.org/10.48550/ARXIV.1210.5984} {Amber: Automatic
  supervision for multi-attribute extraction}.

\bibitem[{Gulhane et~al.(2011)Gulhane, Madaan, Mehta, Ramamirtham, Rastogi,
  Satpal, Sengamedu, Tengli, and Tiwari}]{gulhane2011vertex}
Pankaj Gulhane, Amit Madaan, Rupesh Mehta, Jeyashankher Ramamirtham, Rajeev
  Rastogi, Sandeep Satpal, Srinivasan~H Sengamedu, Ashwin Tengli, and Charu
  Tiwari. 2011.
\newblock \href {https://doi.org/10.1109/ICDE.2011.5767842} {Web-scale
  information extraction with vertex}.
\newblock In \emph{2011 IEEE 27th International Conference on Data
  Engineering}, pages 1209--1220.

\bibitem[{Gupta et~al.(2019)Gupta, Kulkarni, Chanda, Rayasam, and
  Lipton}]{gupta2019amazonqa}
Mansi Gupta, Nitish Kulkarni, Raghuveer Chanda, Anirudha Rayasam, and Zachary~C
  Lipton. 2019.
\newblock Amazonqa: A review-based question answering task.
\newblock \emph{arXiv preprint arXiv:1908.04364}.

\bibitem[{Hao et~al.(2011)Hao, Cai, Pang, and Zhang}]{hao2011swde}
Qiang Hao, Rui Cai, Yanwei Pang, and Lei Zhang. 2011.
\newblock \href {https://doi.org/10.1145/2009916.2010020} {From one tree to a
  forest: A unified solution for structured web data extraction}.
\newblock In \emph{Proceedings of the 34th International ACM SIGIR Conference
  on Research and Development in Information Retrieval}, SIGIR '11, page
  775–784, New York, NY, USA. Association for Computing Machinery.

\bibitem[{Herzig et~al.(2020)Herzig, Nowak, M{\"u}ller, Piccinno, and
  Eisenschlos}]{herzig-etal-2020-tapas}
Jonathan Herzig, Pawel~Krzysztof Nowak, Thomas M{\"u}ller, Francesco Piccinno,
  and Julian Eisenschlos. 2020.
\newblock \href {https://doi.org/10.18653/v1/2020.acl-main.398} {{T}a{P}as:
  Weakly supervised table parsing via pre-training}.
\newblock In \emph{Proceedings of the 58th Annual Meeting of the Association
  for Computational Linguistics}, pages 4320--4333, Online. Association for
  Computational Linguistics.

\bibitem[{Hotti et~al.(2021)Hotti, Risuleo, Magureanu, Moradi, and
  Lagergren}]{hotti2021klarna}
Alexandra Hotti, Riccardo~Sven Risuleo, Stefan Magureanu, Aref Moradi, and Jens
  Lagergren. 2021.
\newblock \href {http://arxiv.org/abs/2111.02168} {The klarna product page
  dataset: A realistic benchmark for web representation learning}.

\bibitem[{Hubert and Arabie(1985)}]{Hubert1985ARI}
Lawrence~J. Hubert and Phipps Arabie. 1985.
\newblock Comparing partitions.
\newblock \emph{Journal of Classification}, 2:193--218.

\bibitem[{Kushmerick(1997)}]{kushmerick1997wrapper}
Nicholas Kushmerick. 1997.
\newblock \emph{Wrapper induction for information extraction}.
\newblock University of Washington.

\bibitem[{Leonhardt et~al.(2020)Leonhardt, Anand, and
  Khosla}]{leonhardt2020boilerplate}
Jurek Leonhardt, Avishek Anand, and Megha Khosla. 2020.
\newblock \href {https://doi.org/10.1145/3366424.3383547} {Boilerplate removal
  using a neural sequence labeling model}.
\newblock \emph{Companion Proceedings of the Web Conference 2020}.

\bibitem[{Li et~al.(2021)Li, Xu, Cui, and Wei}]{li2021markuplm}
Junlong Li, Yiheng Xu, Lei Cui, and Furu Wei. 2021.
\newblock \href {http://arxiv.org/abs/2110.08518} {Markuplm: Pre-training of
  text and markup language for visually-rich document understanding}.

\bibitem[{Lin et~al.(2020)Lin, Sheng, Vo, and Tata}]{lin2020freedom}
Bill~Yuchen Lin, Ying Sheng, Nguyen Vo, and Sandeep Tata. 2020.
\newblock \href {https://doi.org/10.1145/3394486.3403153} {Freedom: A
  transferable neural architecture for structured information extraction on web
  documents}.
\newblock \emph{Proceedings of the 26th ACM SIGKDD International Conference on
  Knowledge Discovery \& Data Mining}.

\bibitem[{Linden et~al.(2003)Linden, Smith, and
  York}]{linden2003recommendations}
G.~Linden, B.~Smith, and J.~York. 2003.
\newblock \href {https://doi.org/10.1109/MIC.2003.1167344} {Amazon.com
  recommendations: item-to-item collaborative filtering}.
\newblock \emph{IEEE Internet Computing}, 7(1):76--80.

\bibitem[{Liu et~al.(2019)Liu, Ott, Goyal, Du, Joshi, Chen, Levy, Lewis,
  Zettlemoyer, and Stoyanov}]{roberta}
Yinhan Liu, Myle Ott, Naman Goyal, Jingfei Du, Mandar Joshi, Danqi Chen, Omer
  Levy, Mike Lewis, Luke Zettlemoyer, and Veselin Stoyanov. 2019.
\newblock \href {https://doi.org/10.48550/ARXIV.1907.11692} {Roberta: A
  robustly optimized bert pretraining approach}.

\bibitem[{Lockard et~al.(2020)Lockard, Shiralkar, Dong, and
  Hajishirzi}]{lockard2020zeroshotceres}
Colin Lockard, Prashant Shiralkar, Xin~Luna Dong, and Hannaneh Hajishirzi.
  2020.
\newblock \href {http://arxiv.org/abs/2005.07105} {Zeroshotceres: Zero-shot
  relation extraction from semi-structured webpages}.

\bibitem[{Petrovski et~al.(2016)Petrovski, Primpeli, Meusel, and
  Bizer}]{conf:ecweb:PetrovskiPMB16}
Petar Petrovski, Anna Primpeli, Robert Meusel, and Christian Bizer. 2016.
\newblock \href
  {http://dblp.uni-trier.de/db/conf/ecweb/ecweb2016.html#PetrovskiPMB16} {The
  wdc gold standards for product feature extraction and product matching.}
\newblock In \emph{EC-Web}, volume 278 of \emph{Lecture Notes in Business
  Information Processing}, pages 73--86.

\bibitem[{Pochat et~al.(2019)Pochat, van Goethem, Tajalizadehkhoob,
  Korczyński, and Joosen}]{Pochat2019TrancoAR}
Victor~Le Pochat, Tom van Goethem, Samaneh Tajalizadehkhoob, Maciej
  Korczyński, and Wouter Joosen. 2019.
\newblock Tranco: A research-oriented top sites ranking hardened against
  manipulation.
\newblock \emph{Proceedings 2019 Network and Distributed System Security
  Symposium}.

\bibitem[{Strehl and Ghosh(2002)}]{Strehl2002NMI}
Alexander Strehl and Joydeep Ghosh. 2002.
\newblock Cluster ensembles --- a knowledge reuse framework for combining
  multiple partitions.
\newblock \emph{J. Mach. Learn. Res.}, 3:583--617.

\bibitem[{Vaswani et~al.(2017)Vaswani, Shazeer, Parmar, Uszkoreit, Jones,
  Gomez, Kaiser, and Polosukhin}]{vaswani2017attention}
Ashish Vaswani, Noam Shazeer, Niki Parmar, Jakob Uszkoreit, Llion Jones,
  Aidan~N. Gomez, Lukasz Kaiser, and Illia Polosukhin. 2017.
\newblock \href {http://arxiv.org/abs/1706.03762} {Attention is all you need}.

\bibitem[{Wang et~al.(2021)Wang, Shiralkar, Lockard, Huang, Dong, and
  Jiang}]{wang2021tcn}
Daheng Wang, Prashant Shiralkar, Colin Lockard, Binxuan Huang, Xin~Luna Dong,
  and Meng Jiang. 2021.
\newblock \href {https://doi.org/10.1145/3442381.3450090} {Tcn: Table
  convolutional network for web table interpretation}.
\newblock \emph{Proceedings of the Web Conference 2021}.

\bibitem[{Xie et~al.(2021)Xie, Huang, Liang, Huang, and Xiao}]{xie2021webke}
Chenhao Xie, Wenhao Huang, Jiaqing Liang, Chengsong Huang, and Yanghua Xiao.
  2021.
\newblock \href {https://doi.org/10.1145/3459637.3482491} {Webke: Knowledge
  extraction from semi-structured web with pre-trained markup language model}.
\newblock In \emph{Proceedings of the 30th ACM International Conference on
  Information \& Knowledge Management}, CIKM '21, page 2211–2220, New York,
  NY, USA. Association for Computing Machinery.

\bibitem[{Zhang et~al.(2018)Zhang, Chen, Ai, Yang, and Croft}]{zhang2018saur}
Yongfeng Zhang, Xu~Chen, Qingyao Ai, Liu Yang, and W.~Bruce Croft. 2018.
\newblock \href {https://doi.org/10.1145/3269206.3271776} {Towards
  conversational search and recommendation: System ask, user respond}.
\newblock In \emph{Proceedings of the 27th ACM International Conference on
  Information and Knowledge Management}, CIKM '18, page 177–186, New York,
  NY, USA. Association for Computing Machinery.

\bibitem[{Zheng et~al.(2007)Zheng, Song, Wen, and Wu}]{zheng2007jointwrapper}
Shuyi Zheng, Ruihua Song, Ji-Rong Wen, and Di~Wu. 2007.
\newblock \href {https://doi.org/10.1145/1281192.1281287} {Joint optimization
  of wrapper generation and template detection}.
\newblock In \emph{Proceedings of the 13th ACM SIGKDD International Conference
  on Knowledge Discovery and Data Mining}, KDD '07, page 894–902, New York,
  NY, USA. Association for Computing Machinery.

\bibitem[{Zhou et~al.(2021)Zhou, Sheng, Vo, Edmonds, and
  Tata}]{zhou2021simplified}
Yichao Zhou, Ying Sheng, Nguyen Vo, Nick Edmonds, and Sandeep Tata. 2021.
\newblock \href {http://arxiv.org/abs/2101.02415} {Simplified dom trees for
  transferable attribute extraction from the web}.

\bibitem[{Zhu et~al.(2006)Zhu, Nie, Wen, Zhang, and Ma}]{zhu2006simul}
Jun Zhu, Zaiqing Nie, Ji-Rong Wen, Bo~Zhang, and Wei-Ying Ma. 2006.
\newblock \href {https://doi.org/10.1145/1150402.1150457} {Simultaneous record
  detection and attribute labeling in web data extraction}.
\newblock In \emph{Proceedings of the 12th ACM SIGKDD International Conference
  on Knowledge Discovery and Data Mining}, KDD '06, page 494–503, New York,
  NY, USA. Association for Computing Machinery.

\bibitem[{Zhu et~al.(2015)Zhu, Kiros, Zemel, Salakhutdinov, Urtasun, Torralba,
  and Fidler}]{zhu2015aligning}
Yukun Zhu, Ryan Kiros, Rich Zemel, Ruslan Salakhutdinov, Raquel Urtasun,
  Antonio Torralba, and Sanja Fidler. 2015.
\newblock Aligning books and movies: Towards story-like visual explanations by
  watching movies and reading books.
\newblock In \emph{Proceedings of the IEEE international conference on computer
  vision}, pages 19--27.

\bibitem[{Álvarez et~al.(2008)Álvarez, Pan, Raposo, Bellas, and
  Cacheda}]{alvarez2008}
Manuel Álvarez, Alberto Pan, Juan Raposo, Fernando Bellas, and Fidel Cacheda.
  2008.
\newblock \href {https://doi.org/https://doi.org/10.1016/j.datak.2007.10.002}
  {Extracting lists of data records from semi-structured web pages}.
\newblock \emph{Data \& Knowledge Engineering}, 64(2):491--509.

\end{thebibliography}
\bibliographystyle{acl_natbib}

\appendix

\label{sec:appendix}

\section{Training Details}
We train all models for up to 5 epochs using an AdamW optimizer with a linear warm-up rate of 10\% and decide on the best learning rate based on development set performance. We search over learning rates of: \{5e-6, 5e-5, 5e-4\}. The best learning rate was 5e-5 for both tasks for RoBERTa and attribute extraction for DOM-LM. 5e-6 was the best learning rate for both tasks for MarkupLM and segmentation for DOM-LM.

\section{Additional Tables}
\begin{table}[H]
\centering
\small
\begin{tabular}{lcrr}
\toprule
\textbf{Attribute} & \textbf{Argmax Correct (\%)} \\
\midrule
Product Name & $89.4$ \\
Review Text & $91.8$ \\
Product Link & $16.2$ \\
\bottomrule
\end{tabular}
\caption{Percentage of false negative examples where the model would be correct if the $argmax$ of the attributes was chosen instead of requiring the activation to be above the threshold $0.5$.} 
\label{tab:falsenegative}
\end{table}

\section{Additional Figures}



\begin{figure}[h]
\centering
\includegraphics[width=1\linewidth]{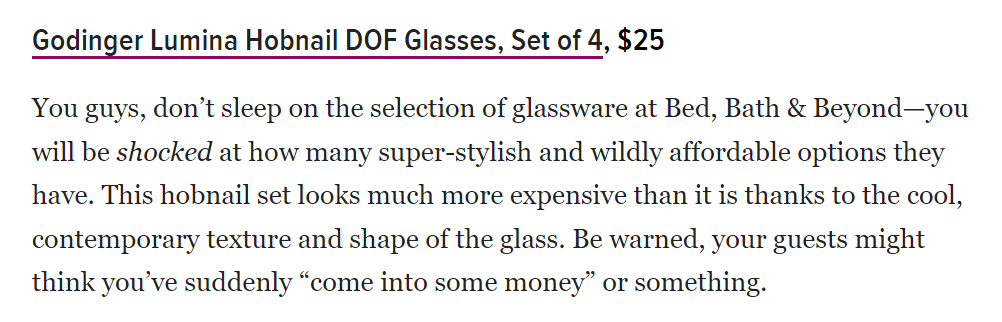}
\caption{An example extraction ``\emph{Godinger Luminal Hobnail Glasses}'' which is both 
a product name and product link.}
\label{fig:multilabelex}
\end{figure}

\begin{figure}[h]
    \centering
    \includegraphics[width=.8\columnwidth]{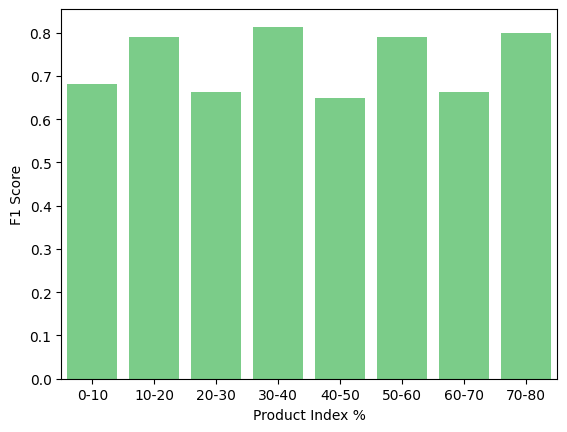}
    \caption{The average attribute extraction \fscore{} based on product index normalized by number of products for the MarkupLM model. The performance is relatively uniform across all indices indicating that product index does not have a significant effect on extraction performance. 
    (We observed a similar trend for both RoBERTa and DOM-LM.)}
    \label{fig:f1productindex}
\end{figure}
\begin{figure}
    \centering
    \includegraphics[width=0.8\columnwidth]{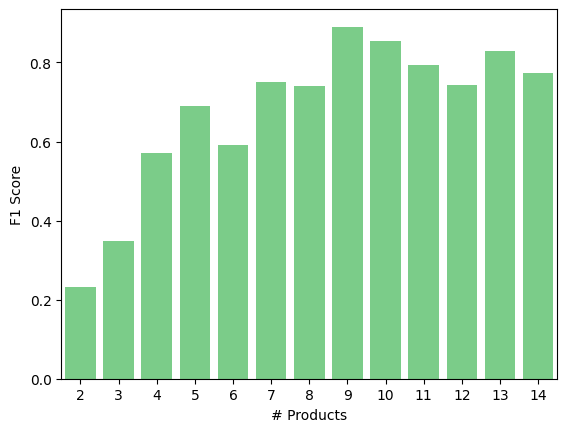}
    \caption{Average attribute extraction \fscore{} based on number of products in a page. In MarkupLM, as the number of products increases, \fscore{} also increases. (We observed a similar trend for both RoBERTa and DOM-LM.)}
    \label{fig:f1numproducts}
\end{figure}

\begin{figure*}
\centering
\includegraphics[width=\textwidth]{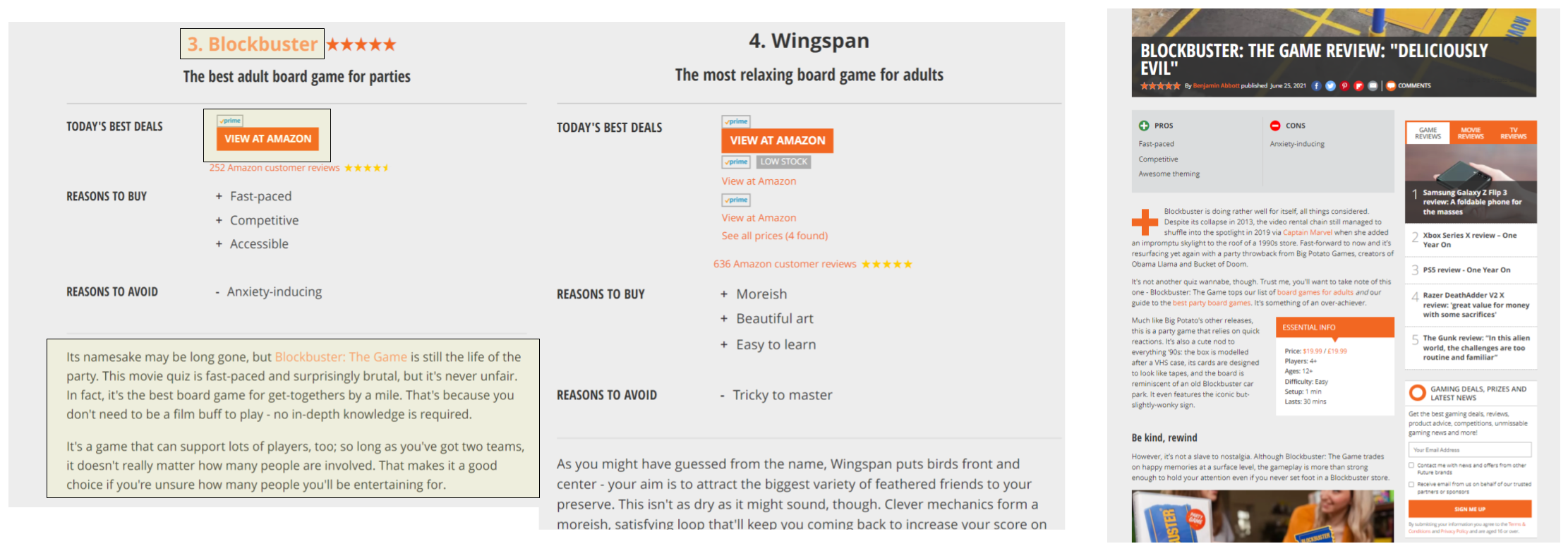}
\caption{An example of a list page compared to a detail page. On the left is a list page with two products: ``Blockbuster'' and ``Wingspan''. Both products have a similar format and directly comparable attributes from our schema: Product Name, Product Link, and Review Text. On the right is a detail page with more in-depth details and longer review text for the ``Blockbuster'' boardgame. (Screenshots from {\tt gamesradar.com})}
\label{fig:listdetailpage}
\end{figure*}


\end{document}